\documentclass[a4paper,fleqn]{cas-sc}
\usepackage[authoryear,longnamesfirst]{natbib}
\usepackage{amsthm}
\usepackage[framemethod=tikz]{mdframed}
\usepackage[ruled,vlined,linesnumbered]{algorithm2e}
\usepackage{amssymb}
\usepackage{amsmath}
\usepackage{graphicx}
\usepackage{float}
\usepackage{xcolor}
\usepackage{wrapfig}
\usepackage{pifont}
\usepackage[center]{caption}
\usepackage[ruled,vlined,linesnumbered]{algorithm2e}

\usepackage{caption}
\let\oldnl\nl% Store \nl in \oldnl
\newcommand{\nonl}{\renewcommand{\nl}{\let\nl\oldnl}}
\def\tsc#1{\csdef{#1}{\textsc{\lowercase{#1}}\xspace}}
\tsc{WGM}
\tsc{QE}
\tsc{EP}
\tsc{PMS}
\tsc{BEC}
\tsc{DE}

\begin{document}
\shorttitle{An Intelligent Mask for Breath-Driven Activity Recognition}
\shortauthors{Sinha et~al.}
\title{\Large{i-Mask: An Intelligent Mask for Breath-Driven Activity Recognition}}

\author[1]{Ashutosh Kumar Sinha}
\ead{ashutosh_2421cs13@iitp.ac.in}
\address[1]{Department of Computer Science and Engineering, Indian Institute of Technology Patna, Bihar, India}

\author[2]{Ayush Patel}
\ead{202318036@daiict.ac.in}
\author[2]{Mitul Dudhat}
\ead{202318024@daiict.ac.in}
\author[2]{Pritam Anand}
\ead{pritam_anand@daiict.ac.in}
\address[2]{Dhirubhai Ambani Institute of Information and Communication Technology (DA-IICT), Gandhinagar, Gujarat, India}

\author[1]{Rahul Mishra}
\ead{rahul_mishra@iitp.ac.in}

\begin{abstract}
The patterns of inhalation and exhalation contain important physiological signals that can be used to anticipate human behavior, health trends, and vital parameters. Human activity recognition (HAR) is fundamentally connected to these vital signs, providing deeper insights into well-being and enabling real-time health monitoring. This work presents i-Mask, a novel HAR approach that leverages exhaled breath patterns captured using a custom-developed mask equipped with integrated sensors. Data collected from volunteers wearing the mask undergoes noise filtering, time-series decomposition, and labeling to train predictive models. Our experimental results validate the effectiveness of the approach, achieving over 95\% accuracy and highlighting its potential in healthcare and fitness applications.
\end{abstract}


\maketitle

\section{Introduction}
Human activity recognition (HAR) has gained significant attention due to its applications in health monitoring, intelligent environments, and human-computer interaction~\cite{10461084,9164991}. 
Traditional HAR approaches employed wearable inertial sensors, vision-based methods, and environmental sensors for HAR. However,  each method has inherent limitations such as discomfort, privacy concerns, or complex deployment requirements~\cite{10614382,10843342}. The human body engages with its environment in diverse ways, one of which is the interaction between the lungs and the external environment through the act of breathing via the nose. The \textbf{breathing pattern} encompasses plenty of useful information that can be processed to fetch different behaviours and health information~\cite{9217864,10478102}. Moreover, the breathing patterns are influenced by metabolic and physiological factors, offering a non-invasive and unobtrusive means of HAR.

The prior literature using masks primarily focuses on physiological monitoring~\cite{r1,r9} and wireless respiratory sensing~\cite{r4,r5}, but lacks AI-driven HAR. While solutions like Masquare~\cite{r1} and FaceBit~\cite{r9} monitor respiration and heart rate, they do not integrate predictive modelling. Similarly, air quality-focused masks~\cite{r2,r3} assess environmental conditions but fail to extract physiological insights. Further, the approaches~\cite{r4,r5} demonstrate remote breath sensing yet struggle with accuracy in dynamic environments. AI-based frameworks~\cite{r6,r7} leverage IoT for health monitoring but overlook breath-based HAR. Furthermore, achieving non-invasive decision-making is still challenging. Thus, it originates in the need for a systematic and structured framework to enable AI-driven physiological analysis in a non-invasive manner to enhance human well-being. 

\begin{figure}[h]
\centering
\includegraphics[width=10.8cm,height=4.5cm]{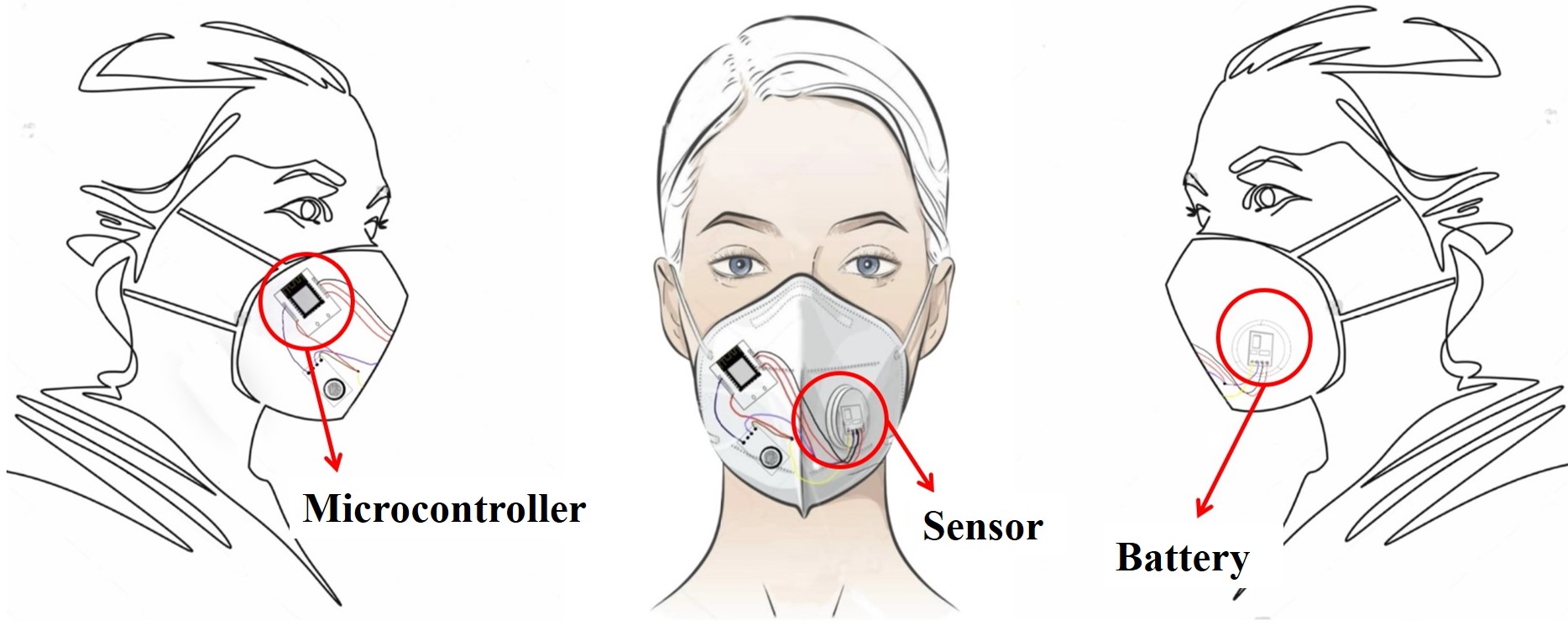}
\caption{A smart mask equipped with a microcontroller, sensor, and battery for real-time monitoring of exhaled air parameters.}
\label{master}
\end{figure}

Inspired by the potential of breath dynamics in HAR derived from prior approaches, we propose \textbf{i-Mask}. This AI-integrated, sensor-equipped bright mask utilizes exhaled breath patterns, precisely temperature and humidity variations, to classify activities. In contrast with the existing HAR approaches, i-Mask offers a \textbf{wearable, non-invasive, and real-time activity classification system}, eliminating the need for external infrastructure. As shown in Fig.~\ref{master}, the proposed system consists of a microcontroller, sensors, and a battery, enabling seamless data acquisition and wireless transmission. The prototype employs compact sensors integrated with the mask to capture inhale-exhale variations continuously. Thus, it provides physiological monitoring with negligible discomfort. 

In doing i-Maks-based analysis, we systematically collect real-world breathing data across different activities. We perform noise filtering, feature extraction, and statistical analysis on the collected data to enhance its quality. Furthermore, we employ learning algorithms to classify activities based on the extracted respiratory patterns. By leveraging AI-driven analysis, i-Mask goes beyond conventional physiological monitoring and enables predictive modelling, setting a new benchmark for breath-based HAR systems.

\noindent \textbf{Contribution:} The contribution is summarized as follows: 

\begin{itemize}
\item We first develop a sensor-integrated wearable mask that detects human activities via temperature and humidity variations and wirelessly transmits data, which also provides preventive measures using an air quality sensor.
\item We next apply noise and vibration filtering to enhance data quality, ensuring robust detection of temperature and humidity peaks. The refined signals undergo time-series decomposition to extract meaningful segments, followed by breath pattern analysis.
\item Finally, we conduct a real-world study to evaluate the effectiveness of the i-Mask system in recognizing human activities based on exhaled breath patterns. The study uses real-time temperature and humidity variations to differentiate locomotion-based activities. Additionally, we benchmark the system’s performance using established classification models and validation metrics, demonstrating its reliability compared to baseline techniques.

\end{itemize}

\noindent \textbf{Roadmap:} Section~\ref{bp} presents related work, preliminaries, and the problem statement. Section~\ref{sec3} details the proposed i-Mask approach, Section~\ref{expr} covers experimental evaluation, and Section~\ref{conc} concludes with future scope.

\section{Background and Preliminaries}\label{bp}

\subsection{Related work}
Prior work in the domain of Smart masks falls in these areas: \textit{physiological monitoring}, \textit{air quality detection}, \textit{wireless AI-based monitoring}, and \textit{AI-driven tracking}.

\begin{table}[H]
\small
\centering
\caption{Summary of Existing Smart Mask Research}
\begin{tabular}{|p{2.5cm}|p{2.5cm}|p{5cm}|p{4cm}|}
\hline
\textbf{Work} & \textbf{System / Work} & \textbf{Key Features} & \textbf{Limitations} \\
\hline
\cite{r1} & \textbf{Masquare} & Cardio-respiratory monitoring via embedded sensors; aesthetic and protective design. & No predictive modeling; focused only on basic health monitoring. \\
\hline
\cite{r9} & \textbf{FaceBit} & Monitors heart rate, respiration, and mask fit using sensors. & Lacks disease prediction and advanced activity recognition. \\
\hline
\cite{r8} & Respiratory Detector & Detects exhaled breath using piezoelectric sounder; works with spirometer; noise robust. & No AI integration or human activity recognition (HAR). \\
\hline
\cite{r2} & Active Mask System & Real-time pathogen exposure reduction using sensors and actuators. & Limited to air quality and pathogen detection only. \\
\hline
\cite{r3} & CO\textsubscript{2}-Sensing Mask & Real-time CO\textsubscript{2} monitoring via opto-chemical sensors. & Cannot assess health impact of CO\textsubscript{2}. \\
\hline
\cite{r4} & WiFi-Respiratory System & Uses Fresnel zone theory for WiFi-based respiratory monitoring. & Accuracy issues in dynamic/multi-human environments. \\
\hline
\cite{r5} & RF Cough Monitor Mask & Wireless, reusable, power-free cough detection using RF transponder. & No disease prediction; RF interference; lacks AI analysis. \\
\hline
\cite{r6} & COVID-Aware System & Combines edge devices, IoT sensors, and DL models for face mask and environment monitoring. & No breath analysis; no HAR integration. \\
\hline
\cite{r7} & Post-COVID Smart Masks & Tracks evolution of smart masks from comfort to health monitoring. & Lacks fine-grained activity detection and real-time AI. \\
\hline
\textbf{Ours} & \textbf{i-Mask (Proposed)} & Predictive modeling using breath patterns; HAR; uses ML and time-series decomposition. & Innovative, integrates disease prediction with HAR. \\
\hline
\end{tabular}
\label{tab:smartmask-summary}
\end{table}

\subsubsection{\textbf{Physiological Monitoring}}
The authors in~\cite{r1} proposed \textbf{Masquare}, a smart mask for real-time cardio-respiratory monitoring via embedded sensors in air-filtering masks. Masquare preserves mask aesthetics and ensures protection and physiological monitoring for diverse health applications. It emphasizes primarily health monitoring but lacks predictive modelling capabilities. Similarly, the authors in~\cite{r9} presented \textbf{FaceBit}, a smart face mask research platform. It is capable of monitoring heart rate, respiration rate, and mask fit using signals. However, its focus remains on basic physiological monitoring rather than disease prediction and advanced activity recognition. Further, \cite{r8} introduced a piezoelectric sounder-based respiratory detector from exhaled breath. It works well with the medical spirometer and exhibits noise robustness in a variety of settings but does not integrate HAR with AI-driven health monitoring.   

\subsubsection{\textbf{Air Quality and Pathogen Detection}} The authors in~\cite{r2} proposed an active mask system integrating sensors and actuators to detect and reduce airborne pathogen exposure in real-time. The developed prototype utilized a particulate matter sensor for air quality assessment and a piezoelectric actuator to neutralize airborne threats. Despite its effectiveness, the approach limits its applicability, confining it to pathogen detection. Similarly, the authors in~\cite{r3} introduced a facemask integrated with an opto-chemical sensor. It is capable of real-time CO$_2$ monitoring, but cannot analyze its impact on health.  

\subsubsection{\textbf{Wireless and AI-Integrated Monitoring}} 
The study in~\cite{r4} explored WiFi-based respiratory monitoring systems for consumer electronics. It leverages the Fresnel zone theory to enhance signal propagation. The approach still faces challenges in maintaining accuracy across dynamic environments with multiple human subjects. Using an RF harmonic transponder, \cite{r5} suggested a breathable, reusable, and power-free face mask for wireless cough monitoring. Although it works well, its drawbacks include the inability to anticipate specific diseases, vulnerability to radio-frequency interference, and the lack of AI-driven analysis.

\subsubsection{\textbf{AI-Driven Monitoring}}
The authors in~\cite{r6} proposed a system to mitigate the effects of COVID-19 by monitoring environmental conditions. The approach utilizes edge devices, IoT sensors, and deep learning models trained on a face mask detection dataset. However, it lacks breath analysis for health monitoring and does not incorporate HAR. 
Another study in~\cite{r7} examined a number of smart masks that surfaced after the pandemic, emphasizing how they progressed from breathing comfort devices to health monitoring devices. However, it lacks fine-grained activity detection, real-time AI-driven insights, and disease prediction.\\ 

\noindent \textbf{Motivation:}
This work builds on previous studies by introducing i-Mask, a novel approach that uses exhaled breath patterns for HAR. Unlike Masquare \cite{r1}, which focuses on health monitoring through wearable textiles, i-Mask emphasizes predictive modelling of vital signs using time-series decomposition and machine learning. While prior research by \cite{r2}, focuses on pathogen detection, i-Mask extends smart mask capabilities to include physiological insights.

Previous approaches like the one by \cite{r5} primarily focused on cough monitoring, while i-Mask integrates breath pattern analysis. The innovation lies in its ability to predict health conditions through real-time monitoring of exhaled breath patterns, achieving high accuracy in disease prediction, similar to the findings in respiratory monitoring~\cite{r8}. Thus, i-Mask presents a comprehensive, non-invasive wearable solution for both activity recognition and disease prediction.

\begin{figure*}[h]
\centering
\includegraphics[scale=0.075]{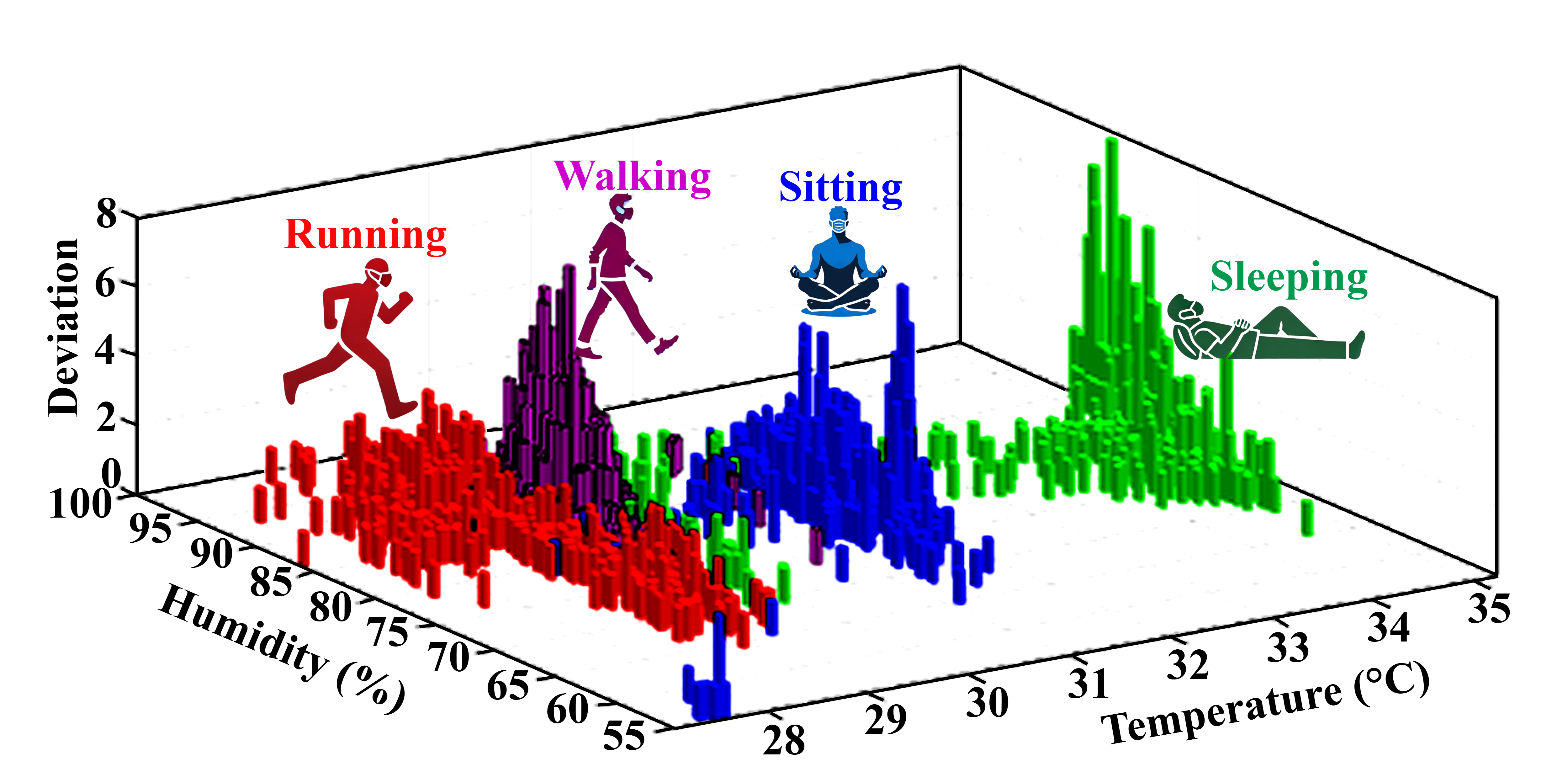}
\caption{An illustration of humidity and temperature variations during exhalation across different activities (running, walking, sitting, and sleeping (\textit{i.e.,} laying down)) while wearing a mask.}
\label{vari}
\end{figure*}

\begin{figure*}
    \centering
    \includegraphics[width=1.0\linewidth]{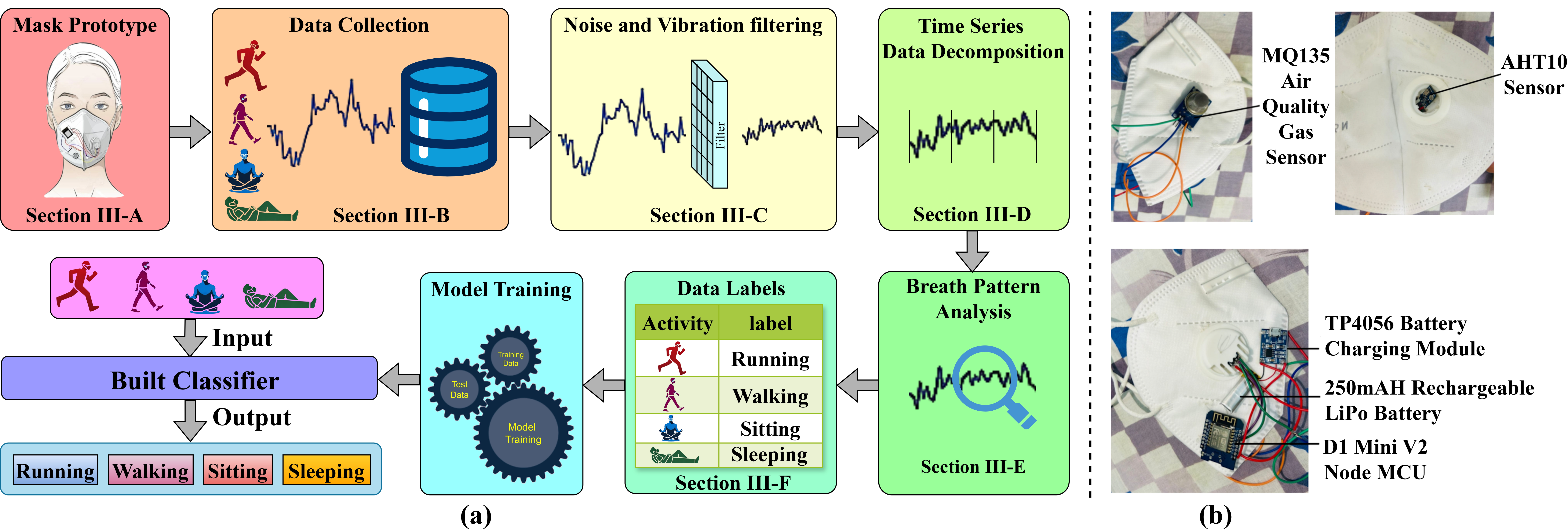}
    \caption{Workflow and Mask prototype. a) The i-Mask system, an exhale breath-based HAR framework, features a smart mask for data acquisition, preprocessing, feature extraction, and activity prediction.  
b) The mask prototype includes an MQ135 gas sensor, AHT10 sensor, LiPo battery, and NodeMCU (microcontroller) for real-time monitoring.}
    \label{overview}
\end{figure*}

\subsection{Physical Insights of Mask Induced Temperature and Humidity Variations}
The i-Mask system leverages the abrupt variations in humidity generated during human inhalation and exhalation. These changes are further complemented by a noticeable rise in temperature~\cite{paeng2024rapid}. The fluctuations in humidity and temperature exhibit distinct patterns depending on various human activities such as running, walking, sitting, and sleeping. The body naturally observes and regulates these variations through mechanisms like chills to maintain homeostasis. While earlier studies have attempted to identify individual inhalation and exhalation patterns to extract vital signatures, they often relied on specialized sensors for data collection and analysis. In contrast, i-Mask employs temperature and humidity sensors to monitor and classify human activities. Each activity produces a unique inhalation and exhalation pattern, leading to corresponding changes in temperature and humidity, as illustrated in Fig.~\ref{vari}. It shows changes in humidity and temperature can affect various human activities, such as walking, running, sitting, and sleeping (\textit{i.e.,} laying down). Temperature (${^o}$C) is represented by the y-axis, humidity (\%) by the x-axis, and deviation levels are displayed by the z-axis. With more dynamic activities (like walking and running) displaying bigger deviations than stationary ones (like sitting and sleeping), the distribution illustrates how contextual influences affect activity recognition.

\subsection{Problem Statement}
The fundamental challenges encountered while developing a prototype mask for real-time monitoring of exhale patterns for HAR are:  \textit{1) how to extract meaningful insights from temperature and humidity data for HAR?, 2) how to handle external noise and vibration while measuring sensory values?, 3) how to perform effective segmentation of the time series data from the developed mask and how to label such data?}

This work investigates and solves the problem of \textit{how can we perform HAR by leveraging exhale breath patterns from mask-based humidity and temperature data with robust feature extraction, noise resilience, and accurate segmentation?}

\section{\textbf{\underline{i-Mask}}: Analyzing Breadth Patterns with an \underline{i}ntelligent \underline{Mask}}\label{sec3}
This work introduces \textbf{i-Mask}, a novel approach for HAR based on exhaled breath patterns. The system captures humidity and temperature variations during exhalation and correlates them with specific human activities, including running, walking, sitting, and sleeping (lying down). The collected data undergoes noise and vibration filtering, followed by exploratory data analysis and time-series segmentation. A machine learning-based classifier is then trained to accurately recognize and classify different activities based on the patterns. The overview of the proposed i-Mask system is depicted in Fig.~\ref{overview} (a). The developed i-Mask supports a non-invasive, real-time activity recognition system based on respiratory signals.

It starts with designing and developing a Mask prototype that incorporates multiple sensors and micro-controller capable of wireless data transmission (Section~\ref{sec3.1}). This mask captures exhale patterns, including variations in humidity and temperature across different activities. The data signals from the mask are recorded and stored for further processing (Section~\ref{sec3.2}). The recorded data undergoes noise and vibration filtering to address noise and vibration that can disrupt temperature and humidity peaks (Section~\ref{sec3.3}). The refined signals are then subjected to time series data decomposition (Section~\ref{sec3.4}) to extract meaningful segments. Following this, breath pattern analysis (Section~\ref{sec3.5}) identifies key respiratory features and then activities are labelled in Section~\ref{sec3.6} for supervised training, linking specific breathing patterns to predefined activity categories.  Algorithm~\ref{algo1} summarizes the process of the i-Mask approach. 

\subsection{Temperature- and Humidity-Induced Activity Recognition Wearable Mask Prototype}\label{sec3.1}
The proposed \textbf{i-Mask} operates on the principle that \textit{variations in exhaled breath temperature and humidity} correspond to different human activities, as illustrated in Fig.~\ref{vari}. 
A well-structured mechanism is required to capture these variations for real-time and precise data acquisition. To achieve this, we developed a \textit{wearable mask prototype to collect and analyze} exhaled breath patterns efficiently.   

Initially, we integrate a \textit{flow sensor} to measure breath dynamics with invasive placement in the nasal passage. This approach seems to create discomfort; thus, we further investigate to reveal a \textit{strong correlation between exhaled breath temperature, humidity changes}, and specific activities while wearing a mask. Such investigation helps to design and develop a sensor-embedded mask that leverages the variations for HAR. During the development phase, we initially tested sensors such as \textit{DHT22} and \textit{BMP180}, which are widely used for environmental monitoring. However, to achieve higher precision and extended functionality, we integrated the \textbf{AHT10} sensor for accurate temperature and humidity measurement. Additionally, we incorporated the \textit{MQ135} sensor to detect variations in air composition, including gases like ammonia, benzene, and smoke, providing valuable contextual data for activity recognition.

Further, the collected sensor data is wirelessly transmitted to a connected smartphone for processing and visualization. To facilitate seamless data collection and transmission, we integrated the D1 Mini V2 NodeMCU, powered by the ESP8266 microcontroller. This compact yet powerful IoT module features built-in Wi-Fi, enabling real-time wireless communication with minimal power consumption, making it an ideal choice for wearable applications. Fig.~\ref{overview} (b) depicts the different components of the developed mask prototype.

To ensure optimal sensor placement and accurate readings, the AHT10 sensor was positioned inside the mask, close to the nose and mouth, allowing precise capture of exhaled breath temperature and humidity. MQ135 sensor and a mini NodeMCU were mounted externally to avoid interference from direct airflow while ensuring seamless connectivity. The sensors were wired to the NodeMCU, which acted as the central processing unit, transmitting real-time data via \textbf{Wi-Fi} to a logging system. A rechargeable battery and charger are integrated into the design to maintain uninterrupted operation.

Let $\mathcal{A} = \{a_1, a_2, a_3, \dots, a_p\}$ represent the set of $p$ activity states monitored by analysing exhaled breath pattern data collected via the mask. The temporal variations in exhaled temperature $T_e$ and humidity $H_e$ are modeled as functions of the activity state $a_i$, \textit{i.e.},
\begin{equation}
T_e=f(a_i) + \zeta_t, \quad  H_e = g(a_i) + \zeta_h, \quad \forall i \in {1\le i\le p},
\end{equation}
where $\zeta_t$ and $\zeta_h$ are the disturbances due to noise and environment. This work classifies activities based on the rate of change and statistical variations in $f(a_i)$ and $g(a_i)$ for each activity $a_i$, as it exhibits distinct exhalation signatures. This rate of change is expressed as: 
\begin{equation}
\frac{dT_e}{dt} = s_t a_i + \zeta_t, \quad \frac{dH_e}{dt} = s_h A + \zeta_H,
\end{equation}
$s_t$ and $s_h$ are the scaling factors that indicate how sensitive temperature and humidity changes are to activity intensity. 

Further, the collected sensor data is wirelessly sent to the connected smartphone for further processing and visualization. The expression for real-time data transmission ($D(t)$ at time $t$) is modeled as follows:  
$D(t) = S(T_e, H_e, AQI) + \xi$, where $S(T_e, H_e, AQI)$ is the total number of sensor readings, and $\xi$ is the data loss and transmission delay.

\subsection{Data Collection and Preprocessing}\label{sec3.2}
The exhale breath pattern dataset collection starts after the successful development of the mask prototype with several field trials. In order to perform the data collection, twenty volunteers, each between the ages of 20 and 30,  participated in four different activities while wearing the sensor-equipped mask: \textit{running, walking, sitting, and sleeping (\textit{i.e.,} laying down)}. We chose these activities to examine the effects of physical activity on exhaled breath parameters to reflect different levels of respiratory effort. \textit{All the volunteers gave their consent to recording the breath pattern data.} 

To facilitate real-time data transmission, the volunteers linked the \textit{NodeMCU} to their \textit{mobile hotspots}, enabling uninterrupted Wi-Fi-based data logging. This setup minimizes data loss and ensures reliable collection. The sampling rate of the data collection is set to one second to perform precise temporal monitoring. The activity sessions are kept to \textit{30 minutes} to provide a comprehensive dataset for analysis. We also record the ambient temperature and humidity before each session to create baseline reference points. The volunteers collect data during the day under controlled indoor conditions to reduce external variability. It helps to achieve consistency 
across trials and provides meaningful information about all the activities.

\subsubsection{Time Intervals for Data Collection}
We employ two different time intervals to examine short- and long-term variations in the exhale breath pattern. Firstly, the data samples of temperature and humidity are averaged over the interval of $30$ minutes to accommodate broader changes. The breath cycles are also counted manually to establish a link between peak variations and physiological activity in each session. Next, the data was collected at one-second intervals to examine rapid changes in exhaled temperature and humidity. Such granularity helps to determine the minute physiological variations that cannot be captured in the long run. Thus, we capture the various activities that affect the properties of exhaled breaths by combining short- and long-term features.

\subsubsection{Iterative Data Collection Process}
During data collection, we employ several sessions to increase statistical robustness, reduce variability, and minimize other external factors. To avoid fatigue-based bias, data collection is appropriately scheduled. For instance, volunteers perform sitting and lying down (or sleeping) on \textit{Day 1}, while running and walking on \textit{Day 2}, and so on. Such scheduling minimizes the impact of fatigue due to high-intensity tasks over low-intensity.

\subsubsection{Analysis of Temperature and Humidity Patterns Across Activities}
\label{subsec:temp_humid_analysis}

To quantify the variation in exhaled breath across different activities, we computed key statistical parameters—\textbf{mean} ($\mu$), \textbf{standard deviation} ($\sigma$), and \textbf{range} ($R$)—for both temperature and humidity readings. These were calculated using the following expressions:

\begin{itemize}
    \item \textbf{Mean:} $\mu = \frac{1}{n} \sum_{i=1}^{n} x_i$ — provides the central tendency of the data.
    \item \textbf{Standard Deviation:} $\sigma = \sqrt{ \frac{1}{n} \sum_{i=1}^{n} (x_i - \mu)^2 }$ — quantifies the dispersion or variability of readings.
    \item \textbf{Range:} $R = \max(x) - \min(x)$ — highlights the span between the lowest and highest values.
\end{itemize}

\textbf{Significance:} These statistics serve as foundational descriptors for understanding how physical activity affects breath characteristics. For instance, the mean ($\mu$) gives an idea of the expected temperature or humidity during a given activity, which is useful for profiling. The standard deviation ($\sigma$) reveals how consistent the breathing patterns are—higher values indicate more fluctuation in breath signals, often linked to intense activity. The range captures the full extent of variations, helping identify extreme conditions or outliers.

\vspace{0.5em}

Table~\ref{tab:temperature} presents the temperature statistics. We observe that low-intensity activities like \textit{sitting} and \textit{sleeping} yield higher average temperatures ($\mu > 31^\circ$C), likely due to slower, deeper breathing and limited airflow exchange. In contrast, during \textit{running} and \textit{walking}, increased ventilation and shorter breath cycles result in lower temperature values ($\mu \approx 29.5^\circ$C to $30.1^\circ$C).

Table~\ref{tab:humidity} highlights that humidity is notably higher during active states like \textit{running} and \textit{walking} ($\mu > 73\%$), driven by greater moisture release from elevated respiratory rates. \textit{Sitting} shows the lowest mean humidity, likely due to steady and shallow breathing patterns.

The standard deviation is also insightful—\textit{running} displays the highest $\sigma$ in both temperature and humidity, reflecting variability in breathing patterns due to physical stress. Meanwhile, \textit{sleeping} shows the lowest $\sigma$, suggesting highly regular breathing.

These statistical measures not only describe physiological differences across activity levels but also serve as effective features for downstream classification models. They enable a non-invasive and continuous understanding of physical activity using breath-based biosignals. In Section~\ref{sec3.6}, we utilize these features for constructing machine learning models that classify activities based on breath temperature and humidity.

\vspace{1em}

\begin{table}[H]
    \centering
    \captionsetup{justification=centering, labelsep=colon}
    \caption{Temperature variations across different activities}
    \label{tab:temperature}
    \begin{tabular}{|l|c|c|c|}
        \hline
        \textbf{Activity} & \textbf{Mean ($^{\circ}$C)} & \textbf{Std. Dev.} & \textbf{Range ($^{\circ}$C)} \\ \hline
        Running & 29.5 & 1.2 & 28.3 -- 30.7 \\ \hline
        Walking & 30.1 & 1.1 & 28.9 -- 31.2 \\ \hline
        Sitting & 31.8 & 0.9 & 30.7 -- 32.9 \\ \hline
        Sleeping & 32.3 & 0.8 & 31.4 -- 33.1 \\ \hline
    \end{tabular}
\end{table}

\vspace{1em}

\begin{table}[H]
    \centering
    \captionsetup{justification=centering, labelsep=colon}
    \caption{Humidity variations across different activities}
    \label{tab:humidity}
    \begin{tabular}{|l|c|c|c|}
        \hline
        \textbf{Activity} & \textbf{Mean (\%)} & \textbf{Std. Dev.} & \textbf{Range (\%)} \\ \hline
        Running & 75.2 & 2.1 & 72.0 -- 78.5 \\ \hline
        Walking & 73.5 & 1.8 & 71.2 -- 76.3 \\ \hline
        Sitting & 68.9 & 1.5 & 67.4 -- 70.5 \\ \hline
        Sleeping & 71.2 & 1.7 & 69.5 -- 73.8 \\ \hline
    \end{tabular}
\end{table}

\subsubsection{Thresholding and Imputation for Outlier and Missing Data Handling}

Before performing any in-depth analysis, it is crucial to ensure that the collected data is clean, consistent, and reliable. In our study, we observed that raw sensor readings occasionally exhibited abrupt spikes or drops, which were not consistent with the expected physiological or environmental behavior during the recorded activities. These anomalies were typically caused by brief sensor malfunctions, user movements that momentarily obstructed readings, or external environmental disturbances.

To handle such issues, we adopted a simple yet effective threshold-based method for outlier detection. This approach is advantageous because it is easy to implement, computationally inexpensive, and provides transparent logic for what constitutes "normal" versus "abnormal" readings—making it ideal for both real-time systems and offline data analysis pipelines.

We calculated thresholds individually for each activity and each variable (temperature and humidity). The idea was to define a safe range of expected sensor values for each activity, based on observed minimum and maximum values in our dataset. To allow for natural variation and minor sensor noise, we added a small tolerance margin, denoted as $\delta_x$, to these observed extremes. The final lower and upper bounds for any activity $a$ and variable $x$ are calculated using:

\begin{equation}
    L_a = \min(\text{range}_a) - \delta_x, \qquad U_a = \max(\text{range}_a) + \delta_x
\end{equation}

For temperature, we used a tolerance of $\delta_x = 0.3\,^{\circ}\mathrm{C}$, and for humidity, $\delta_x = 1.1\%$. These values were selected based on the typical fluctuations seen in environmental sensors and were confirmed through exploratory data analysis.

\vspace{1em}

\noindent
\textit{For example}, for the \textbf{Running} activity:
\begin{itemize}
    \item The observed minimum and maximum temperature were 28.3°C and 30.7°C, respectively. Applying the temperature tolerance gives:
    \[
    L_{\text{Running}}^{\text{temp}} = 28.3 - 0.3 = 28.0,\quad U_{\text{Running}}^{\text{temp}} = 30.7 + 0.3 = 31.0
    \]
    \item The observed minimum and maximum humidity values were 72.1\% and 78.4\%. Applying the humidity tolerance:
    \[
    L_{\text{Running}}^{\text{humidity}} = 72.1 - 1.1 = 71.0,\quad U_{\text{Running}}^{\text{humidity}} = 78.4 + 1.1 = 79.5
    \]
\end{itemize}

\noindent
Any value falling outside these bounds was flagged as an outlier and removed from the dataset. This step significantly enhanced the quality of the data and ensured that downstream tasks—such as activity recognition and trend analysis—were not affected by transient errors or sensor drift.

Our goal with this method was not only to clean the data but also to do so in a way that is explainable. Unlike more complex anomaly detection techniques such as Isolation Forests or neural autoencoders, our method provides explicit thresholds that can be validated and audited by domain experts. This transparency is particularly important in health and behavioral analytics, where explainability often matters as much as accuracy.

\vspace{1em}

\noindent
A summary of the final bounds used for outlier detection is presented in Table~\ref{tab:calculated_thresholds}.

\begin{table}[H]
\centering
\caption{Computed lower and upper bounds for temperature ($^{\circ}$C) and humidity (\%) across different activities for outlier detection.}
\label{tab:calculated_thresholds}
\begin{tabular}{|l|cc|cc|}
\hline
\multirow{2}{*}{\textbf{Activity}} & \multicolumn{2}{c|}{\textbf{Temperature ($^{\circ}$C)}} & \multicolumn{2}{c|}{\textbf{Humidity (\%)}} \\
\cline{2-5}
                                   & Lower Bound & Upper Bound & Lower Bound & Upper Bound \\
\hline
Running    & 28.0 & 31.0 & 71.0 & 79.5 \\
Walking    & 28.6 & 31.5 & 70.2 & 77.3 \\
Sitting    & 30.4 & 33.2 & 66.4 & 71.5 \\
Sleeping   & 31.1 & 33.4 & 68.5 & 74.8 \\
\hline
\end{tabular}
\end{table}

\textbf{Missing Value Imputation.}  
To handle missing entries—either due to sensor silence or removal of outliers—we applied linear interpolation. Given two valid values at $t_{k-1}$ and $t_{k+1}$, the missing value at $t_k$ was estimated as:

\[
x(t_k) = x(t_{k-1}) + \frac{x(t_{k+1}) - x(t_{k-1})}{t_{k+1} - t_{k-1}} \cdot (t_k - t_{k-1})
\]

This approach preserved temporal continuity and enabled downstream models to process uninterrupted sequences.

The simplicity and transparency of both thresholding and imputation ensured explainability and efficiency, which are critical in health and behavioral sensing applications.

\subsection{Noise and Vibration Filtering}\label{sec3.3}
\begin{figure}[h]
\centering
\hspace{-0.5cm}\includegraphics[scale=0.4]{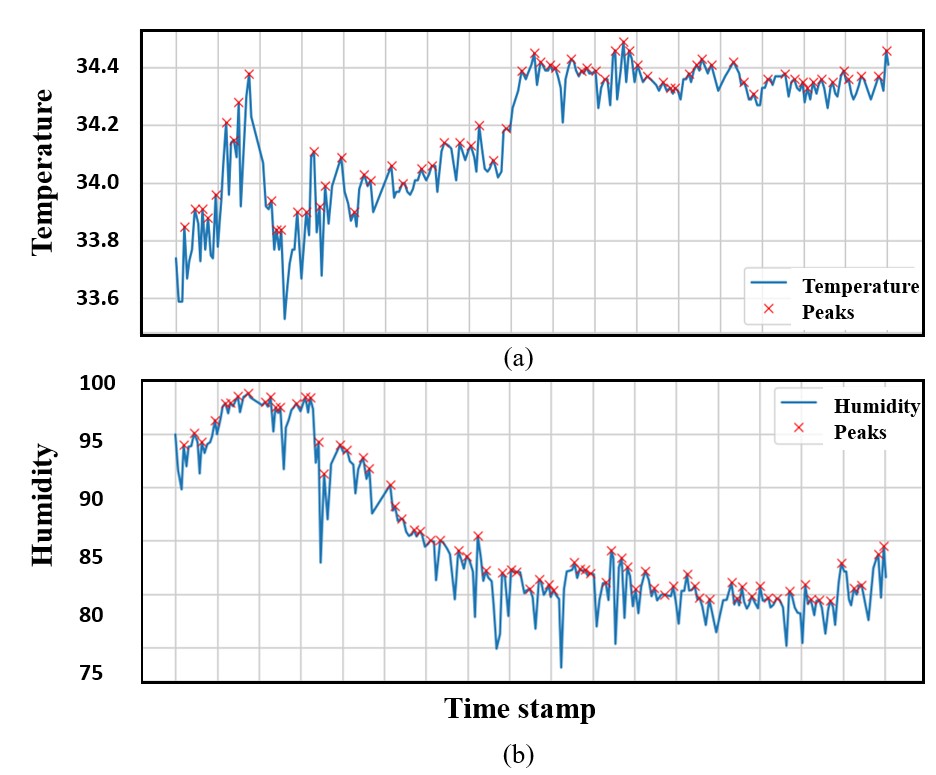}
\caption{Peaks in \textit{(a) temperature} and \textit{(b) humidity} signals over time, chosen from the collected data.}
\label{temp}
\end{figure}

We apply a set of software filters to highlight the change in the temperature and humidity peaks after the successful acquisition of data.  
Such application is similar to those established in the previous methods~\cite{4122029, 10570893, flohr}. Fig~\ref{temp} depicts the detection of peaks in temperature and humidity over a considered time period from the collected data. The peaks show significant local maxima, which is essential for examining respiration patterns. This peak detection method extracts the important respiratory cycle properties by recognizing inhalation and exhalation events based on changes in temperature and humidity patterns.  

We first apply a low-pass filter at $400$ Hz to remove excess high-frequency noise in the collected data. Next, we employ a wavelet transform to correlate the temperature and humidity with a series of wavelet templates, highlighting impulsive changes associated with respiration cycles. The wavelet transform ($\mathcal{WT}$) of a signal $\mathcal{X}(t)$ is expressed as follows:
\begin{equation}
\mathcal{WT}(\alpha, \beta) = \int_{-\infty}^{\infty} \mathcal{X}(t) \Psi^{\ast}_{(a, b)}(t) dt,
\end{equation}
where $\Psi_{(a,b)}$ is the wavelet function, scaled by $a$ and translated by $b$, and $\ast$ is the complex conjugate. The wavelet filter lessens environmental variables, such as slight changes in airflow or other sources, which are often more oscillatory than impulsive. Heuristically, the wavelet filter is adjusted to match the sensor's reaction to inhaling and exhaling patterns. In this work, we employ a first-order Gaussian derivative wavelet function defined as follows:
\begin{equation}
\Psi(t) = -\frac{t}{\sigma^2} e^{-\frac{t^2}{2\sigma^2}}.
\end{equation}

To roughly represent the changes brought on by breathing, each breathing cycle appears as a rounded impulse modelled as a Gaussian bump as follows:
\begin{equation}
g(t) = e^{-\frac{(t - \mu)^2}{2\sigma^2}},
\end{equation}
where $\mu$ denotes the mean inhalation/exhalation time and $\sigma$  controls the spread of the Gaussian. The structures' elastic response to these changes differentiates the original signal. A wide-band wavelet filter is obtained by summing the responses after correlating with ten linearly spaced scales of this wavelet with center frequencies ranging from $7$ to $54$ Hz. We estimate such frequencies using respiration-induced vibrations. The material and sensitivity of the sensors influence the frequencies. 

The wavelet filter alone cannot prepare the sensory signal for peak identification since breathing fluctuations vary among people and environments. We apply an envelope detector to the wavelet-filtered signal to combine these waves into a single peak. The envelope $E(t)$ is computed using the Hilbert transform $H(x)$:
\begin{equation}
E(t) = \sqrt{x^2(t) + H^2(x(t))}
\end{equation}
where $H(x(t))$ represents $x(t)$'s Hilbert transform. The final low-pass stage of the envelope detector adjusts to $1.5$ times the typical maximum adult respiratory rate, which makes it possible to record extremely rapid breathing patterns precisely.

\subsubsection{Scaling and Synchronizing Sensor Data}
\label{subsec:scaling_sync}

To ensure consistent and meaningful analysis across different sensor modalities (e.g., temperature and humidity), two essential preprocessing steps are employed: \textbf{min-max scaling} and \textbf{time alignment}. These steps help standardize the data and ensure that it is temporally coherent before performing statistical analysis or feeding it into machine learning models.

\vspace{1mm}
\noindent\textbf{Min-Max Scaling:} Sensor readings can vary widely in scale. For instance, humidity may range from 65\% to 80\%, whereas temperature may vary from 28°C to 34°C. If such features are fed directly into a model, those with larger magnitudes may disproportionately influence the output. To avoid this, we normalize each feature to a common scale, typically between 0 and 1, using min-max scaling. The formula is:

\[
x_{\text{scaled}} = \frac{x - x_{\min}}{x_{\max} - x_{\min}}
\]

As an example, suppose we have a temperature value of 30.5°C. Given that the minimum and maximum temperature readings in the dataset are 28.0°C and 34.0°C respectively, the scaled value becomes:

\[
x_{\text{scaled}} = \frac{30.5 - 28.0}{34.0 - 28.0} = \frac{2.5}{6} \approx 0.417
\]

Thus, 30.5°C is transformed into approximately 0.417 in the normalized range. This transformation ensures that both temperature and humidity are comparable and balanced during training or analysis.

\vspace{2mm}
\noindent\textbf{Time Alignment and Synchronization:}  
Even when both temperature and humidity are recorded from a single sensor, slight timing discrepancies can still occur due to internal buffering, software timestamping delays, or variation in logging intervals. These delays, though minimal, may accumulate and affect time-dependent analysis such as feature windowing or time-series modeling. Thus, synchronization remains a critical preprocessing step to ensure data consistency across the timeline.

To achieve this, we first define a uniform sampling frequency \( f_s \) (e.g., \( f_s = 1 \text{ Hz} \)), which determines the expected time interval \( \Delta t = \frac{1}{f_s} \). We then round all observed timestamps \( t_i \) to the nearest multiple of \( \Delta t \), such that each time point aligns with a discrete set \( \{ t_0, t_1, t_2, \dots \} \). When gaps appear in the data after rounding, we estimate the missing values using interpolation. If \( x(t) \) is a signal with missing value at time \( t_k \), and surrounding known values at \( t_{k-1} \) and \( t_{k+1} \), then the interpolated value is computed as:
\[
x(t_k) = x(t_{k-1}) + \frac{x(t_{k+1}) - x(t_{k-1})}{t_{k+1} - t_{k-1}} \cdot (t_k - t_{k-1})
\]
This approach ensures that each data point aligns with a consistent and evenly spaced timeline, enabling robust downstream processing. The resulting synchronized dataset allows for precise temporal comparisons and feature extraction, forming a stable basis for statistical analysis or machine learning tasks.

\vspace{2mm}
\noindent Together, these two steps—scaling and synchronization—prepare raw sensor data for robust and interpretable analysis, enabling downstream algorithms to operate more effectively across diverse activity states.

\subsection{Time Series Data Decomposition}\label{sec3.4}
We decompose the time series data into constituent components to further analyze the respiratory patterns captured through temperature and humidity sensors. This decomposition facilitates the isolation of underlying trends, periodicity, and noise, thereby enhancing the clarity and interpretability of the respiratory signals. This work employs the Seasonal-Trend Decomposition using the Loess (STL)~\cite{sanchez2012using} method for this purpose. STL is a robust and flexible decomposition technique that separates a time series into \textit{three additive components} to uncover hidden patterns and physiological correlations:  

\noindent \textbf{Trend} ($T_t$): The trend component captures the long-term shifts in temperature and humidity. It reflects the environmental changes or physiological responses (\textit{e.g.,} increased body heat). 

\noindent \textbf{Seasonality} ($S_t$): This component detects regular variations associated with breathing cycles, such as temperature and humidity rising with expiration and falling with inhalation.

\noindent \textbf{Residuals} ($R_t$): It highlights sensor noise or possible anomalies by representing random changes after eliminating trend and seasonality. 
Let $y_t$ be the observed data at time $t$, then the decomposition following an additive model is: $y_t = T_t + S_t + R_t$.

\begin{figure}[h]
\centering
\includegraphics[width=0.7\textwidth]{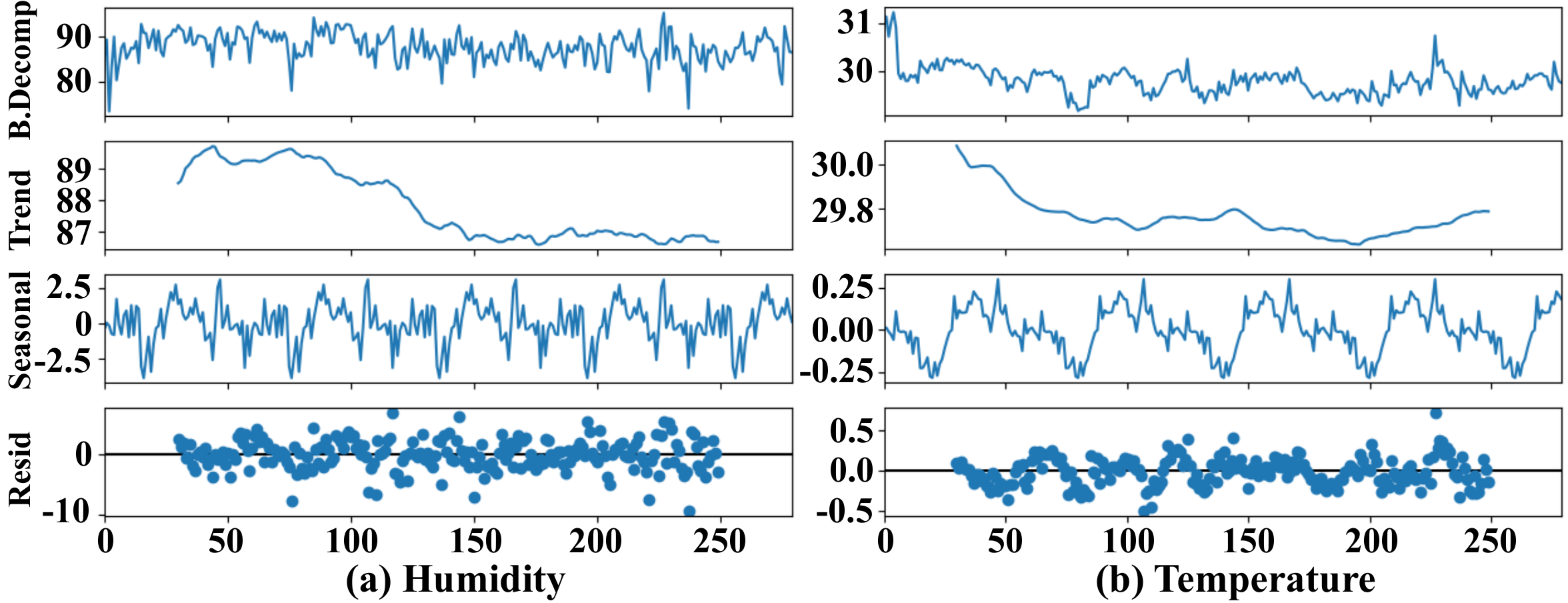}
\caption{Extracting Trend, Seasonality, and Residual Variations from the STL Decomposition of Temperature and Humidity Time Series. Resd. = Residual and B.Decomp. = Before Decomposition.}
\label{decomp}
\end{figure}

STL decomposition and peak analysis offer essential information on the interaction of temperature, humidity, and breathing patterns. Fig.~\ref{decomp} depicts the breakdown of temperature and humidity time series into three components: (a) \textit{Trend} (showing long-term fluctuations); (b) \textit{Seasonality} (depicting periodic fluctuations caused by recurrent environmental patterns); and (c) \textit{Residuals} (\textit{i.e.,} short-term noise and irregularities). By separating these elements, the analysis improves our comprehension of environmental dynamics and makes it easier to identify anomalies, predict their occurrence, and investigate any possible physiological implications.

\subsection{Exhaled Breath Pattern Analysis}\label{sec3.5}
This section covers a thorough Exploratory Data Analysis (EDA) of the temperature and humidity data (discussed Section~\ref{sec3.2}) to find trends in breath dynamics. We systematically analyze Breath Pattern Peaks in Section~\ref{sec3.51}, explore Correlation Trends in Section~\ref{sec3.52} and examine the Statistical Distribution of Data in Section~\ref{sec3.53}.

\subsubsection{\textbf{Breathing Patterns Peak Analysis}}~\label{sec3.51}
The next step following the time series data decomposition is determining the peaks linked to the breath cycle. Such peaks represent the 
inhale and exhale phases, depicting cyclical changes in temperature and humidity. Here, we aim to learn more about the rhythmic nature of breath patterns and their relation to environmental influences by investigating these oscillations. 

We employ the following steps for peak detection in time series data:
\begin{itemize}
\item We first employ a peak detection algorithm to identify local minima in the seasonal component w.r.t. the duration of inhale and exhale. This helps in examining the peaks' matched breadth cycles, their height and spacings.

\item Next, we compare and validate the detected peaks with the manually recorded breathing cycles per session. It helps validate the link between breath cycle, temperature, and humidity.

\item  We also collected data on multiple breath cycles over various time intervals and corrected them with changes in temperature and humidity. It helps establish the breathing patterns' interaction with environmental factors and validates the periodicity. 
 \end{itemize}

\begin{figure}
\centering
\includegraphics[width=.42\textwidth]{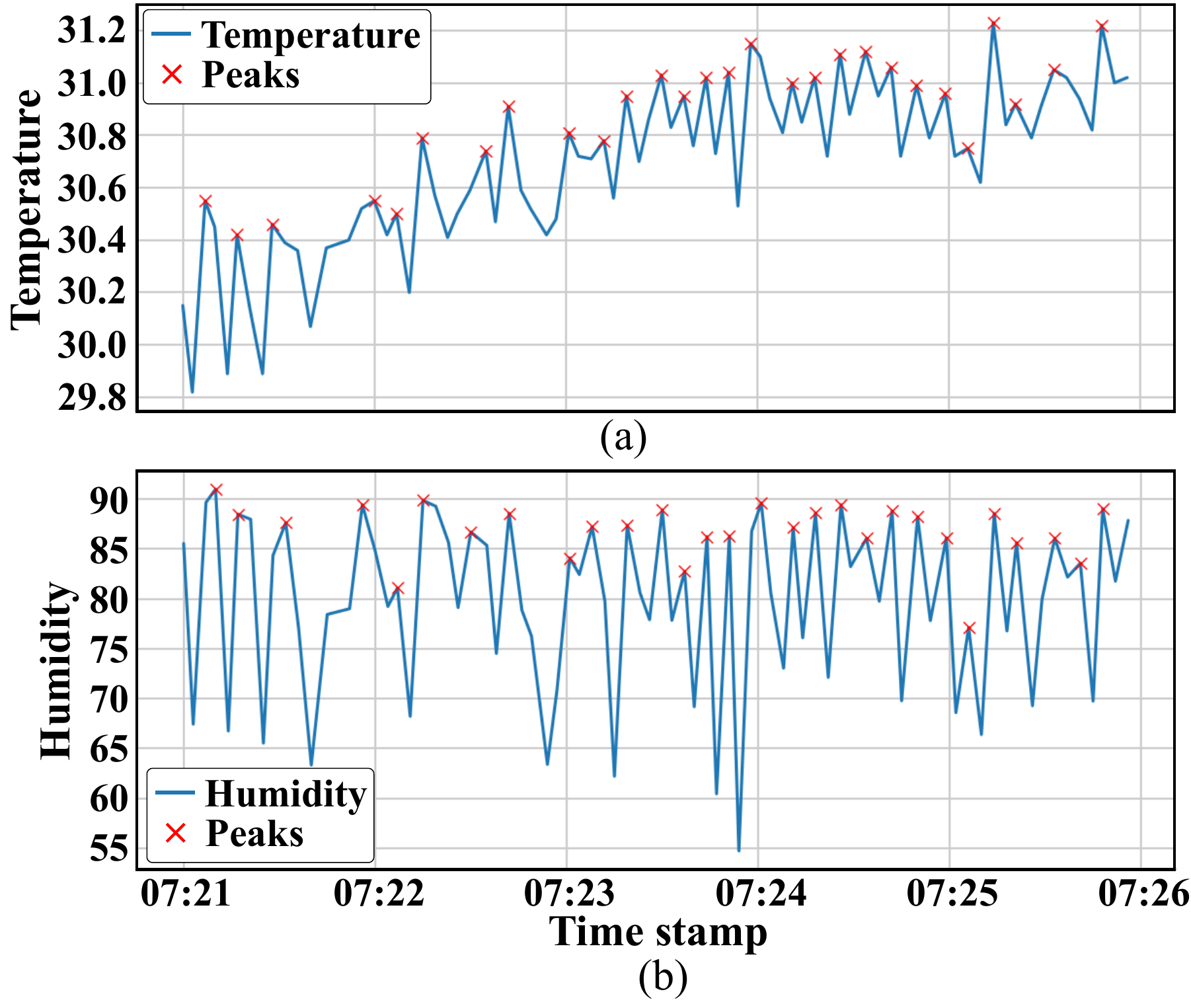}
\caption{Peak detection in (a) temperature and (b) humidity variations during breathing cycles.}
\label{hum}
\end{figure}

\textit{Visualizing Peak Data:}  
Figs.~\ref{temp} and \ref{hum} show the peak detection results for various time stamps. We depict the identified peaks next to the unprocessed temperature and humidity data to confirm that they matched the anticipated breathing patterns. The analysis involves the following steps:  

$\bullet$ To ensure each peak matched a breath, we compared the frequency of detected peaks with manually recorded breathing cycles. We next modify the detection settings after identifying unevenly spaced peaks.

$\bullet$ To determine the effect of the environment and physical effort that affected temperature and humidity changes, the stability of peak detection is examined by a range of activities.  

\begin{figure*}
\centering
\includegraphics[scale=0.2]{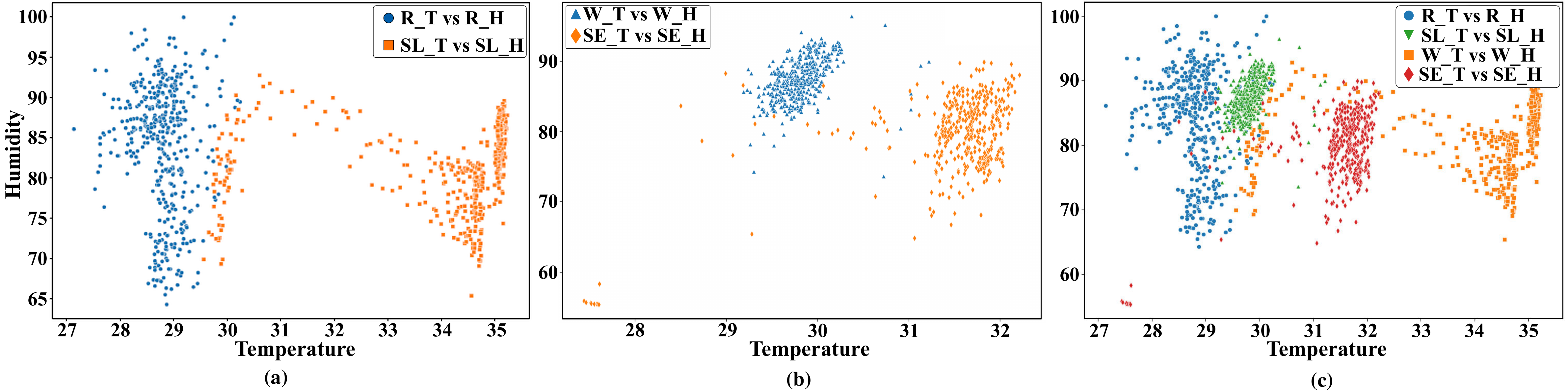}
\caption{Scatter Plot Representation of Temperature vs. Humidity Across Different Activities}
\label{corr1}
\end{figure*}

\subsubsection{\textbf{Correlation Analysis}}~\label{sec3.52} 
\begin{figure}[h]
\centering
\includegraphics[width=.49\textwidth]{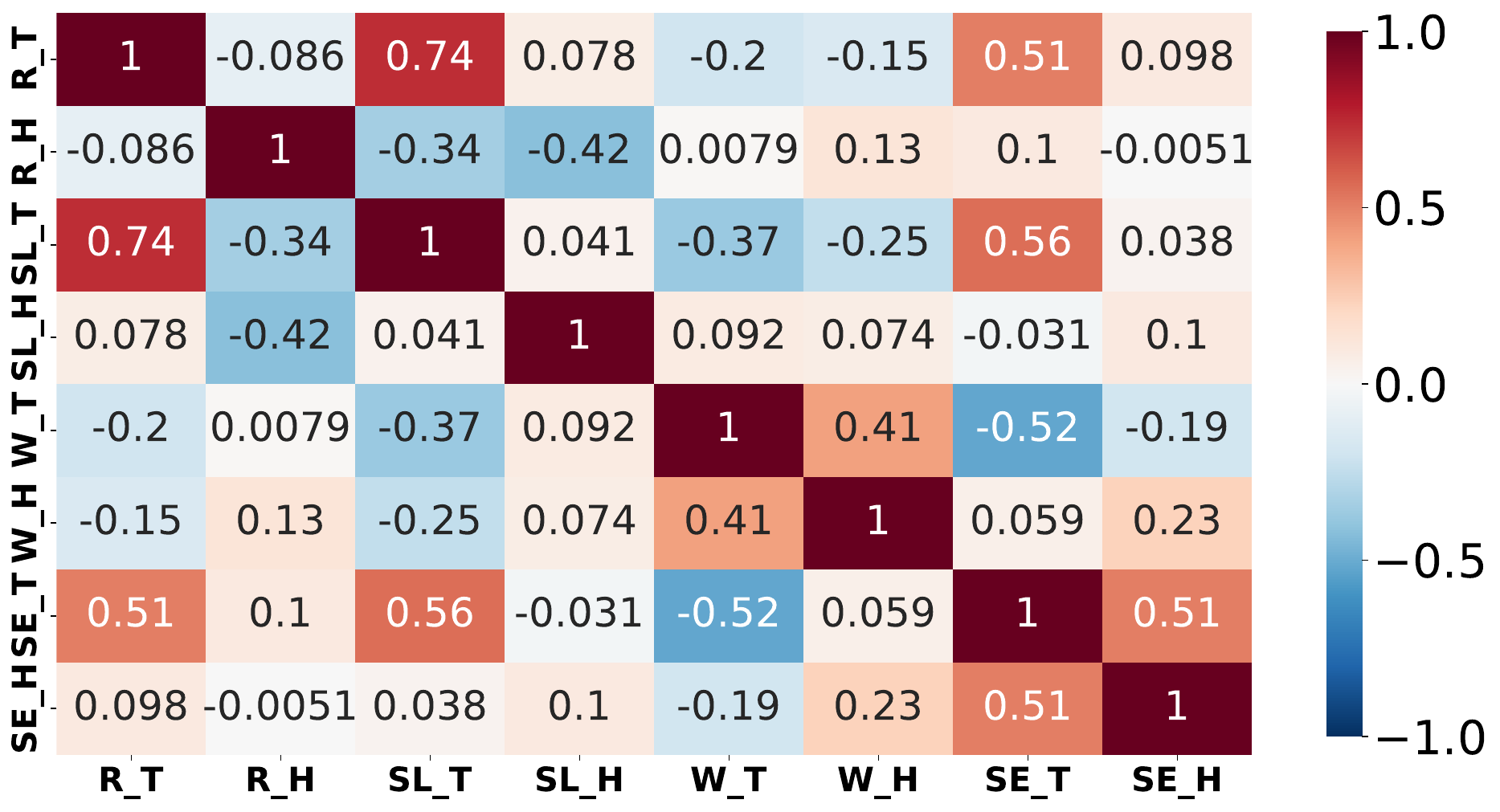}
\caption{Correlation between temperature and humidity across different conditions (R\_T: Running Temperature, R\_H: Running Humidity, SL\_T: Sleeping Temperature, SL\_H: Sleeping Humidity, SE\_T: Sitting Temperature, SE\_H: Sitting Humidity, W\_T: Walking Temperature, W\_H: Walking Humidity).}
\label{heat}
\end{figure}

We next present a comprehensive \textbf{correlation analysis} between temperature and humidity across different activities, focusing on both positive and negative relationships, as depicted in Fig.~\ref{heat}. \textbf{Positive correlations} reveal that Running Temperature (\textit{R\_T}) is moderately positively correlated with Sitting Humidity (\textit{SE\_H}) and Walking Humidity (\textit{W\_H}), suggesting that increased temperature during running leads to higher humidity levels during sitting and walking. Additionally, a positive correlation exists between \textit{SE\_H} and \textit{W\_H}, although it is weaker than the one with \textit{R\_T}. 

On the other hand, \textbf{negative correlations} highlight that \textit{R\_T} has a moderate negative relationship with both Sitting Temperature (\textit{SE\_T}) and Sleeping Temperature (\textit{SL\_T}), indicating that as the running temperature rises, the temperatures during sitting and sleeping tend to decrease. A similar inverse relationship is observed between \textit{SE\_T} and \textit{SL\_T}, though it is weaker compared to their correlation with \textit{R\_T}. Furthermore, \textbf{strong correlations} reveal a significant negative correlation between \textit{R\_T} and \textit{SL\_T}, as well as between \textit{R\_T} and \textit{SE\_H} and Sleeping Humidity (\textit{SL\_H}), illustrating the inverse relationship between running temperature and these variables. Finally, \textbf{weak correlations} were observed between \textit{SE\_T} and Walking Temperature (\textit{W\_T}), \textit{R\_H} and \textit{W\_H}, and \textit{SE\_T} and \textit{SE\_H}, showing slight tendencies for these variables to increase together. This analysis provides valuable insights into how environmental factors like temperature and humidity fluctuate in response to different activities and their interplay. Similarly, Fig.~\ref{corr1} shows the temperature and humidity of the exhaled breath for four activities: running, sleeping, sitting, and walking.

\subsubsection{Data Distribution}~\label{sec3.53}
Data distribution analysis in activity recognition using breath patterns is essential to the EDA. It identifies the presence of \textit{outliers} in some distributions and suggests occasional deviations due to sensor fluctuations, external environmental factors, or anomalous activity conditions. Thus, it contributes to more accurate \textit{sensor calibration, anomaly detection, and personalized environmental modelling}.

\begin{figure}[h]
\centering
\includegraphics[width=0.7\textwidth]{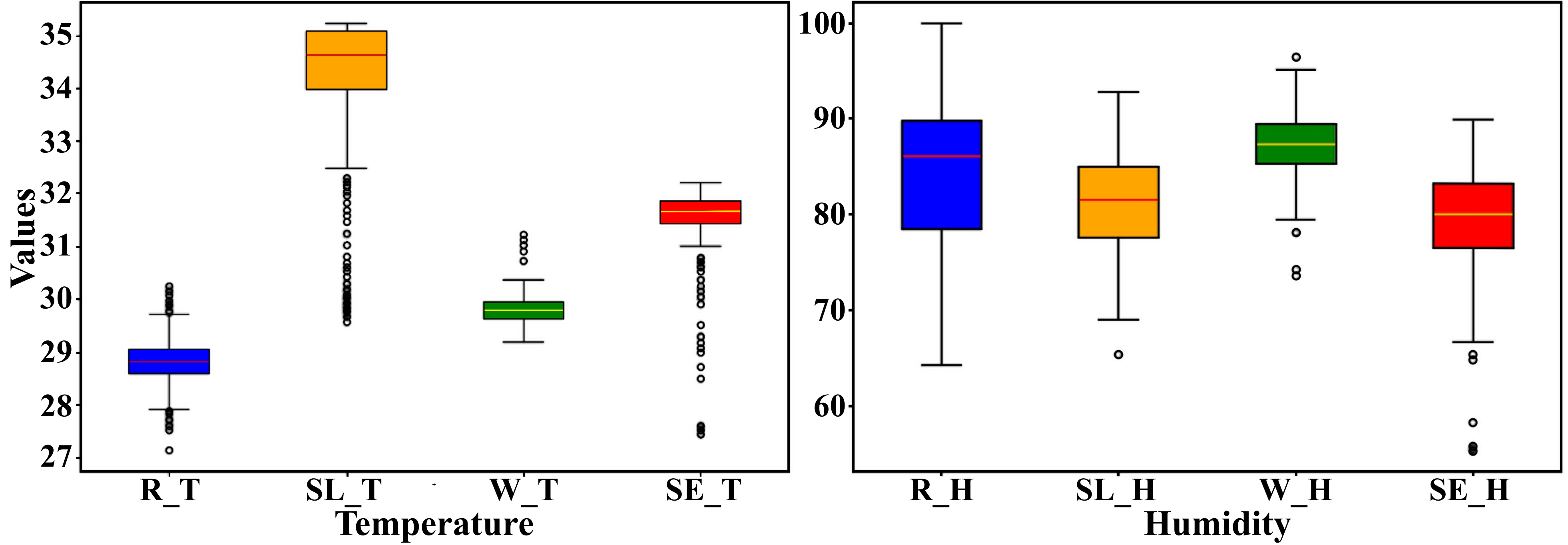}
\caption{Data distribution of Temperature and Humidity Distributions while performing various activities.}
\label{heat1}
\end{figure}

Using box plots, Fig.~\ref {heat1} visualizes temperature and humidity measurement distribution across different activities. The central tendency of temperature across activities suggests different thermal conditions influenced by movement and posture. The higher metabolic heat generation during sitting and sleeping, which have relatively steady distributions, gives activities like walking and running a more extensive spread. Furthermore, variations in humidity levels among activities may reflect variations in breathing, sweating, and environmental exposure. 
Higher humidity variations in active states (e.g., running and walking) indicate increased moisture levels, while more controlled conditions (e.g., sleeping and sitting) result in a narrower interquartile range.

\subsection{Data Labelling}\label{sec3.6}
We initially performed manual data labelling, with domain experts assigning activity labels. Though the process of manual labelling is very accurate for limited data, it is time-consuming and subject to influence due to human bias. Thus, we can incorporate an AI-based labelling technique to enhance efficiency by automating annotations.  

Next, we compare the automated and manual labels, where the results depict a high correlation, signifying the correctness of automated labelling. Thus, combining manual verification with automated labelling, i-Mask ensures accurate, scalable, and efficient data annotation.

\subsection{Training and Testing}
Using the collected sensor data and labeled activities from the previous step, the i-Mask approach applies machine learning models for activity prediction. Algorithms such as Decision Trees (DT), k-Nearest Neighbors (kNN), and Support Vector Machines (SVM) are trained to classify activities based on temperature and humidity variations.  

During training, extracted features from the labeled dataset were used to optimize model performance. The models were then tested using a stratified k-fold cross-validation approach to ensure robustness and generalizability. The final trained model was selected based on its accuracy in correctly predicting activities across different conditions.

\begin{algorithm}[t]
\caption{\textbf{Activity Recognition from Exhaled Breath Patterns}}
\label{algo1}
\smallskip
\nonl \noindent \textbf{Step 1:} \textit{Mask Prototype Development}\\ 
\smallskip
Design a mask using sensors (AHT10, MQ135, BMP180)\;  
Ensure real-time data transmission via ESP8266 NodeMCU\;  
\smallskip
\nonl \noindent \textbf{Step 2:} \textit{Data Collection}\\ 
\smallskip
Volunteers perform activities\;  
Record temperature and humidity variations\;  
Store ambient conditions for baseline reference\; 
\smallskip
\nonl \noindent \textbf{Step 3:} \textit{Noise and Vibration Filtering}\\
\smallskip
Apply low-pass filter to remove high-frequency noise\;  
Use wavelet transform for peak detection in exhalation\;  
Employ Hilbert transform for envelope extraction\;
\smallskip
\nonl \noindent \textbf{Step 4:} \textit{Time Synchronization, Scaling and Outlier Handling}\\
\smallskip
Align timestamps and interpolate missing values using linear interpolation\;  
Normalize signals to range [0, 1] via min-max scaling to remove magnitude bias\;  
Detect and remove outliers using activity-specific thresholds with tolerance margins\;
\smallskip
\nonl \noindent \textbf{Step 5:} \textit{Time-Series Data Decomposition}\\
\smallskip
Decompose signals using Seasonal-Trend decomposition (STL)\;  
Analyze periodic trends and remove low-confidence anomalies\; 
\smallskip
\nonl \noindent \textbf{Step 6:} \textit{Breath Pattern Analysis and Labeling}\\
\smallskip
Identify exhalation peaks aligned with respiratory cycles\;  
Manually label activity windows based on characteristic breath profiles\;  
Validate and refine annotations with AI-assisted suggestions\;
\smallskip
\nonl \noindent \textbf{Step 7:} \textit{Feature Extraction and Model Training}\\
\smallskip
Extract statistical and temporal features (mean, std, peak distance, etc.)\;  
Train classification models (e.g., SVM, Random Forest, LSTM)\;  
Apply stratified k-fold cross-validation for robustness\;  
Select optimal model based on validation accuracy and consistency\;
\smallskip
\nonl \noindent \textbf{Step 8:} \textit{Activity Recognition}\\
\smallskip
Deploy trained model for real-time activity classification\;  
Continuously update model with new data for adaptive learning\;  
\end{algorithm}

\section{Experiments}\label{expr}
This section discusses the experimental findings and analysis of the proposed i-Mask approach using different classification models and settings. 

\subsection{Experimental Setup}
We evaluate the performance of the proposed i-Mask approach using real-world data collected from participants performing different activities while wearing the sensor-equipped mask. The dataset comprises the humidity and temperature variations captured across four activities: \textit{Running, Walking, Sitting, and Sleeping (laying down)}. We perform multiple trials in controlled indoor conditions during the experiment to ensure robustness and reproducibility.  

The ESP8266 NodeMCU microcontroller facilitates real-time data transmission, and all collected data was logged at a 1-second sampling rate. Each activity session lasted 30 minutes, and to reduce bias, data collection was performed on different days for high- and low-intensity activities.

\subsection{Data Preprocessing and Feature Engineering}
We employ the following preprocessing steps to prepare the dataset for model training and analysis:  a) a low-pass filter at 400 Hz is applied to remove high-frequency noise; b) sensor data is transformed to highlight periodic breathing patterns; c) Min-max scaling is applied to bring temperature and humidity readings to a uniform scale; and d) the dataset was divided into meaningful time windows to preserve activity transitions.

\subsection{Activity Classification Model Training}
For activity prediction, four machine learning models were trained and evaluated: Decision Tree (DT), which provides rule-based classification with high interpretability; k-nearest Neighbors (kNN), a distance-based model that requires feature standardization; Random Forest (RF), an ensemble learning approach that enhances generalization; and Support Vector Machine (SVM), which identifies optimal decision boundaries in high-dimensional space. We partitioned the dataset into 80\% training and 20\% testing sets to ensure an unbiased evaluation.

\noindent $\bullet$ \textbf{Hyper-parameter Optimization:}
We apply GridSearchCV for hyper-parameter tuning across different models to improve accuracy. The optimal depth and node split criterion is determined for the decision tree, with entropy selected as the best criterion. In the k-nearest neighbours model, the best k-value is k=3, ensuring optimal neighbourhood size for classification. Further, the random forest model performs best using 100 trees, enhancing stability and generalization. Finally, we tested different kernels for the support vector machine and obtained that the RBF kernel was the most effective.

\subsection{Performance Evaluation}
We evaluate the trained model using standard classification metrics, including accuracy, precision, recall, and F1-score (Table~\ref{tab:model_performance}). These metrics provide a holistic view of the model’s ability to correctly classify activities based on the recorded temperature and humidity variations..

\begin{table}[htbp]
    \centering
    \caption{Performance Metrics for different models.}
    \begin{tabular}{|l|c|c|c|c|}
        \hline
        \textbf{Metric} & \textbf{Decision Tree} & \textbf{kNN} & \textbf{Random Forest} & \textbf{SVM} \\
        \hline
        Accuracy & 0.957 & \textbf{0.964} & 0.948 & 0.744 \\
         Precision & 0.958 & \textbf{0.965} & 0.949 & 0.799 \\
        Recall & 0.957 & \textbf{0.964} & 0.948 & 0.744 \\
        F1-score & 0.957 & \textbf{0.964} & 0.948 & 0.736 \\
        \hline
    \end{tabular}
    \label{tab:model_performance}
\end{table}

\noindent \textbf{Takeaway:} \textit{Among all, kNN achieved the highest accuracy (96.4\%), followed closely by the Decision Tree and Random Forest models, while SVM exhibited the lowest performance.}

\subsubsection{\textbf{Performance of (kNN)}}  
\begin{figure}[h]
    \centering
    \includegraphics[width=0.7\linewidth]{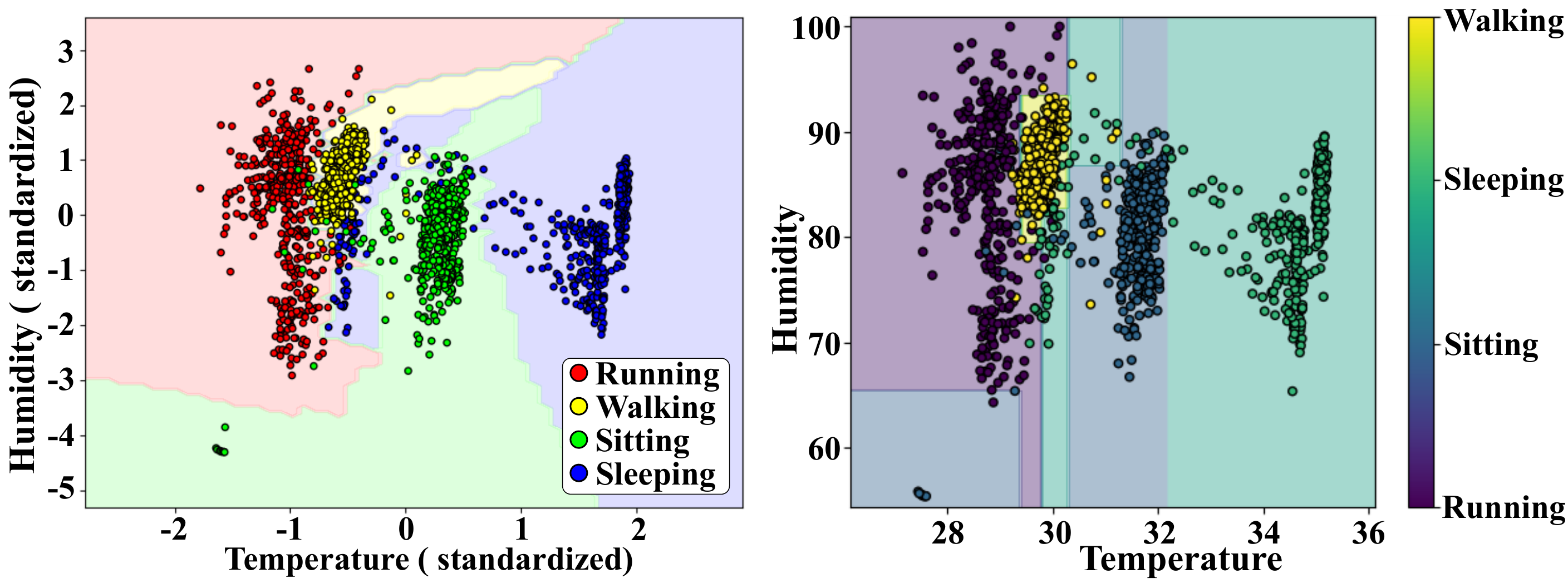}
    \caption{Decision Boundary of kNN Classifier}
    \label{fig:knn_decision_boundary}
\end{figure}

The higher performance outcome of kNN shows that it successfully reflects the underlying patterns of temperature-humidity distribution linked to various activities. This improvement in the performance is due to the distance-based strategy of the classifier that guarantees similar activity states are grouped in feature space. Furthermore, the \textbf{stratified 5-fold cross-validation} technique helped reduce bias and improve model generalization. The confusion matrix (Table~\ref{tab:knn_confusion_matrix}) shows minimal misclassification, particularly in closely related activities such as walking and running, where a few instances were confused due to overlapping breathing patterns. The decision boundary visualization (Figure~\ref{fig:knn_decision_boundary}) further confirms that kNN effectively separates activity states based on feature similarities.

\begin{table}[htbp]
    \centering
    \caption{Confusion Matrix for kNN.}
    \begin{tabular}{|c|c|c|c|c|}
        \hline
        \multicolumn{5}{|c|}{\textbf{Actual / Predicted}} \\
        \hline
        & \textbf{Running} & \textbf{Sitting} & \textbf{Sleeping} & \textbf{Walking} \\
        \hline
        Running & 421 & 0 & 1 & 13 \\
        Sitting & 7 & 424 & 2 & 2 \\
        Sleeping & 3 & 8 & 414 & 10 \\
        Walking & 9 & 4 & 3 & 419 \\
        \hline
    \end{tabular}
    \label{tab:knn_confusion_matrix}
\end{table}

\begin{table}[htbp]
    \centering
    \caption{Classification Report for kNN.}
    \begin{tabular}{|l|c|c|c|c|}
        \hline
        \textbf{Class} & \textbf{Precision} & \textbf{Recall} & \textbf{F1-score} & \textbf{Support} \\
        \hline
        Running & 0.96 & 0.97 & 0.96 & 435 \\
        Sitting & 0.97 & 0.97 & 0.97 & 435 \\
        Sleeping & 0.99 & 0.95 & 0.97 & 435 \\
        Walking & 0.94 & 0.96 & 0.95 & 435 \\
        \hline
        Accuracy &  &  & 0.96 & 1740 \\
        Macro avg & 0.96 & 0.96 & 0.96 & 1740 \\
        Weighted avg & 0.96 & 0.96 & 0.96 & 1740 \\
        \hline
    \end{tabular}
    \label{tab:knn_classification_report}
\end{table}

\subsubsection{\textbf{Decision Tree Classifier}}
The result in Table~\ref{tab:model_performance} demonstrates that the  DT achieved an accuracy of 95.7\%, slightly less than kNN. It is because of the benefits of its rule-based classification, which allows for transparent decision-making. However, we observe overfitting in some cases, particularly in training samples where activities had sharp transitions. The confusion matrix (Table~\ref{tab:dt_confusion_matrix}) reveals that while most activities were classified correctly, there were occasional misclassifications between running and walking, likely due to similarities in exhaled breath dynamics. Despite this, the high precision and recall scores demonstrate that the model can still provide reliable classifications, as depicted in Table~\ref{tab:dt_classification_report}.

\begin{table}[htbp]
    \centering
    \caption{Confusion Matrix for Decision Tree.}
    \begin{tabular}{|c|c|c|c|c|}
        \hline
        \multicolumn{5}{|c|}{\textbf{Actual / Predicted}} \\
        \hline
        & \textbf{Running} & \textbf{Sitting} & \textbf{Sleeping} & \textbf{Walking} \\
        \hline
        Running & 85 & 0 & 2 & 5 \\
        Sitting & 1 & 75 & 1 & 0 \\
        Sleeping & 0 & 0 & 90 & 3 \\
        Walking & 3 & 0 & 0 & 83 \\
        \hline
    \end{tabular}%
    \label{tab:dt_confusion_matrix}
\end{table}

\begin{table}[htbp]
    \centering
    \caption{Classification Report for Decision Tree.}
    \begin{tabular}{|c|c|c|c|c|}
        \hline
        \textbf{Class} & \textbf{Precision} & \textbf{Recall} & \textbf{F1-score} & \textbf{Support} \\
        \hline
        Running & 0.96 & 0.92 & 0.94 & 92 \\
        Sitting & 1.00 & 0.97 & 0.99 & 77 \\
        Sleeping & 0.97 & 0.97 & 0.97 & 93 \\
        Walking & 0.91 & 0.97 & 0.94 & 86 \\
        \hline
        Accuracy &  &  & 0.96 & 348 \\
        Macro avg & 0.96 & 0.96 & 0.96 & 348 \\
        Weighted avg & 0.96 & 0.96 & 0.96 & 348 \\
        \hline
    \end{tabular}
    \label{tab:dt_classification_report}
\end{table}

\subsubsection{\textbf{Performance of Random Forest and SVM}}  
The Random Forest model achieved 94.8\% accuracy, ranking third among classifiers. Its ensemble approach mitigates overfitting by aggregating decision trees, enhancing generalization, but its computational complexity and sensitivity to redundant features slightly lowered its performance compared to Decision Tree and kNN. While effective in classifying stationary activities like sitting and sleeping, it showed some confusion in high-intensity activities like running. In contrast, SVM performed the worst, with 74.4\% accuracy, struggling with high-dimensional feature interactions that hindered its ability to separate overlapping activity states.

\section{Conclusion}\label{conc}
This work introduced i-Mask, a novel human activity recognition (HAR) approach that utilizes exhaled breath patterns captured via a custom-designed bright mask equipped with integrated sensors. Our findings demonstrate the feasibility of non-invasive, real-time activity recognition, with the kNN classifier achieving an accuracy of 96.4\%. The finding of our approach emphasizes the potential of breath-based physiological signals for HAR, with implications for healthcare, fitness, and personalized monitoring. Future work will explore integrating deep learning models to enhance classification accuracy and robustness. We will also expand the capabilities of i-Mask by extending the study using a variety of datasets, integrating multi-sensor fusion, and investigating disease prediction through breath analysis. 

\section{Declaration of competing interest} \label{Declaration of competing interest}
The authors declare that they have no known competing financial interests or personal relationships that could have appeared to influence the work reported in this paper. 

\bibliographystyle{elsarticle-num-names}
\bibliography{Bibliography}

\begin{thebibliography}{20}
\expandafter\ifx\csname natexlab\endcsname\relax\def\natexlab#1{#1}\fi
\providecommand{\url}[1]{\texttt{#1}}
\providecommand{\href}[2]{#2}
\providecommand{\path}[1]{#1}
\providecommand{\DOIprefix}{doi:}
\providecommand{\ArXivprefix}{arXiv:}
\providecommand{\URLprefix}{URL: }
\providecommand{\Pubmedprefix}{pmid:}
\providecommand{\doi}[1]{\href{http://dx.doi.org/#1}{\path{#1}}}
\providecommand{\Pubmed}[1]{\href{pmid:#1}{\path{#1}}}
\providecommand{\bibinfo}[2]{#2}
\ifx\xfnm\relax \def\xfnm[#1]{\unskip,\space#1}\fi
%Type = Article
\bibitem[{Hussain et~al.(2024)Hussain, Khan, Khan, Bhatt, Farouk, Bhola, and
  Baik}]{10461084}
\bibinfo{author}{A.~Hussain}, \bibinfo{author}{S.~U. Khan},
  \bibinfo{author}{N.~Khan}, \bibinfo{author}{M.~W. Bhatt},
  \bibinfo{author}{A.~Farouk}, \bibinfo{author}{J.~Bhola},
  \bibinfo{author}{S.~W. Baik},
\newblock \bibinfo{title}{A hybrid transformer framework for efficient activity
  recognition using consumer electronics},
\newblock \bibinfo{journal}{IEEE Transactions on Consumer Electronics}
  \bibinfo{volume}{70} (\bibinfo{year}{2024}) \bibinfo{pages}{6800--6807}.
%Type = Article
\bibitem[{Mishra et~al.(2022)Mishra, Gupta, Gupta, and Dutta}]{9164991}
\bibinfo{author}{R.~Mishra}, \bibinfo{author}{A.~Gupta}, \bibinfo{author}{H.~P.
  Gupta}, \bibinfo{author}{T.~Dutta},
\newblock \bibinfo{title}{A sensors based deep learning model for unseen
  locomotion mode identification using multiple semantic matrices},
\newblock \bibinfo{journal}{IEEE Transactions on Mobile Computing}
  \bibinfo{volume}{21} (\bibinfo{year}{2022}) \bibinfo{pages}{799--810}.
%Type = Article
\bibitem[{Wang et~al.(2024)Wang, Huang, Zhao, Zhu, Huang, and Wu}]{10614382}
\bibinfo{author}{P.~Wang}, \bibinfo{author}{H.~Huang},
  \bibinfo{author}{L.~Zhao}, \bibinfo{author}{B.~Zhu},
  \bibinfo{author}{H.~Huang}, \bibinfo{author}{H.~Wu},
\newblock \bibinfo{title}{Extre: Extended temporal-spatial network for
  consumer-electronic wifi-based human activity recognition},
\newblock \bibinfo{journal}{IEEE Transactions on Consumer Electronics}
  (\bibinfo{year}{2024}) \bibinfo{pages}{1--1}.
%Type = Article
\bibitem[{Mishra and Gupta(2025)}]{10843342}
\bibinfo{author}{R.~Mishra}, \bibinfo{author}{H.~P. Gupta},
\newblock \bibinfo{title}{Towards understanding the impact of participant and
  its wearable devices in federated learning},
\newblock \bibinfo{journal}{IEEE Transactions on Mobile Computing}
  (\bibinfo{year}{2025}) \bibinfo{pages}{1--13}.
%Type = Inproceedings
\bibitem[{Mongelli et~al.(2020)Mongelli, Orani, Cambiaso, Vaccari, Paglialonga,
  Braido, and Catalano}]{9217864}
\bibinfo{author}{M.~Mongelli}, \bibinfo{author}{V.~Orani},
  \bibinfo{author}{E.~Cambiaso}, \bibinfo{author}{I.~Vaccari},
  \bibinfo{author}{A.~Paglialonga}, \bibinfo{author}{F.~Braido},
  \bibinfo{author}{C.~E. Catalano},
\newblock \bibinfo{title}{Challenges and opportunities of iot and ai in
  pneumology},
\newblock in: \bibinfo{booktitle}{2020 23rd Euromicro Conference on Digital
  System Design (DSD)}, \bibinfo{year}{2020}, pp. \bibinfo{pages}{285--292}.
%Type = Article
\bibitem[{Zhang et~al.(2024)Zhang, Wang, and Li}]{10478102}
\bibinfo{author}{H.~Zhang}, \bibinfo{author}{Q.~Wang}, \bibinfo{author}{N.~Li},
\newblock \bibinfo{title}{Da-net: A dense attention reconstruction network for
  lung electrical impedance tomography (eit)},
\newblock \bibinfo{journal}{IEEE Internet of Things Journal}
  \bibinfo{volume}{11} (\bibinfo{year}{2024}) \bibinfo{pages}{22107--22115}.
%Type = Inproceedings
\bibitem[{Fischer et~al.(2021)Fischer, Wittmann, Baucells~Costa, Zhou, Joost,
  and Lukowicz}]{r1}
\bibinfo{author}{H.~F. Fischer}, \bibinfo{author}{D.~Wittmann},
  \bibinfo{author}{A.~Baucells~Costa}, \bibinfo{author}{B.~Zhou},
  \bibinfo{author}{G.~Joost}, \bibinfo{author}{P.~Lukowicz},
\newblock \bibinfo{title}{Masquare: A functional smart mask design for health
  monitoring},
\newblock in: \bibinfo{booktitle}{Proc. ACM ISWC}, \bibinfo{year}{2021}, p.
  \bibinfo{pages}{175–178}.
%Type = Article
\bibitem[{Curtiss et~al.(2022)Curtiss, Rothrock, Bakar, Arora, Huang,
  Englhardt, Empedrado, Wang, Ahmed, Zhang, Alshurafa, and Hester}]{r9}
\bibinfo{author}{A.~Curtiss}, \bibinfo{author}{B.~Rothrock},
  \bibinfo{author}{A.~Bakar}, \bibinfo{author}{N.~Arora},
  \bibinfo{author}{J.~Huang}, \bibinfo{author}{Z.~Englhardt},
  \bibinfo{author}{A.-P. Empedrado}, \bibinfo{author}{C.~Wang},
  \bibinfo{author}{S.~Ahmed}, \bibinfo{author}{Y.~Zhang},
  \bibinfo{author}{N.~Alshurafa}, \bibinfo{author}{J.~Hester},
\newblock \bibinfo{title}{Facebit: Smart face masks platform},
\newblock \bibinfo{journal}{Proc. ACM Interact. Mob. Wearable Ubiquitous
  Technol.} \bibinfo{volume}{5} (\bibinfo{year}{2022}).
%Type = Article
\bibitem[{Zhang et~al.(2024)Zhang, Bao, Jiahao, and Zhu}]{r4}
\bibinfo{author}{L.~Zhang}, \bibinfo{author}{R.~Bao},
  \bibinfo{author}{C.~Jiahao}, \bibinfo{author}{Y.~Zhu},
\newblock \bibinfo{title}{Smart city healthcare: Non-contact human respiratory
  monitoring with wifi-csi},
\newblock \bibinfo{journal}{IEEE Transactions on Consumer Electronics}
  \bibinfo{volume}{70} (\bibinfo{year}{2024}) \bibinfo{pages}{5960--5968}.
%Type = Article
\bibitem[{Ye et~al.(2022)Ye, Ling, Yang, Xu, Zhu, Yan, and Chen}]{r5}
\bibinfo{author}{Z.~Ye}, \bibinfo{author}{Y.~Ling}, \bibinfo{author}{M.~Yang},
  \bibinfo{author}{Y.~Xu}, \bibinfo{author}{L.~Zhu}, \bibinfo{author}{Z.~Yan},
  \bibinfo{author}{P.-Y. Chen},
\newblock \bibinfo{title}{A breathable, reusable, and zero-power smart face
  mask for wireless cough and mask-wearing monitoring},
\newblock \bibinfo{journal}{ACS Nano} \bibinfo{volume}{16}
  (\bibinfo{year}{2022}) \bibinfo{pages}{5874--5884}.
%Type = Article
\bibitem[{Kalavakonda et~al.(2021)Kalavakonda, Masna, Bhuniaroy, Mandal, and
  Bhunia}]{r2}
\bibinfo{author}{R.~R. Kalavakonda}, \bibinfo{author}{N.~V.~R. Masna},
  \bibinfo{author}{A.~Bhuniaroy}, \bibinfo{author}{S.~Mandal},
  \bibinfo{author}{S.~Bhunia},
\newblock \bibinfo{title}{A smart mask for active defense against coronaviruses
  and other airborne pathogens},
\newblock \bibinfo{journal}{IEEE Consumer Electronics Magazine}
  \bibinfo{volume}{10} (\bibinfo{year}{2021}) \bibinfo{pages}{72--79}.
%Type = Article
\bibitem[{Escobedo et~al.(2022)Escobedo, Fern{\'a}ndez-Ramos, L{\'o}pez-Ruiz,
  Moyano-Rodr{\'\i}guez, Mart{\'\i}nez-Olmos, P{\'e}rez~de Vargas-Sansalvador,
  Carvajal, Capit{\'a}n-Vallvey, and Palma}]{r3}
\bibinfo{author}{P.~Escobedo}, \bibinfo{author}{M.~Fern{\'a}ndez-Ramos},
  \bibinfo{author}{N.~L{\'o}pez-Ruiz},
  \bibinfo{author}{O.~Moyano-Rodr{\'\i}guez},
  \bibinfo{author}{A.~Mart{\'\i}nez-Olmos}, \bibinfo{author}{I.~P{\'e}rez~de
  Vargas-Sansalvador}, \bibinfo{author}{M.~Carvajal},
  \bibinfo{author}{L.~Capit{\'a}n-Vallvey}, \bibinfo{author}{A.~Palma},
\newblock \bibinfo{title}{Smart facemask for wireless co2 monitoring},
\newblock \bibinfo{journal}{Nature Communications} \bibinfo{volume}{13}
  (\bibinfo{year}{2022}) \bibinfo{pages}{72}.
%Type = Article
\bibitem[{Aydemir and Arslan(2023)}]{r6}
\bibinfo{author}{F.~Aydemir}, \bibinfo{author}{S.~Arslan},
\newblock \bibinfo{title}{A system design with deep learning and iot to ensure
  education continuity for post-covid},
\newblock \bibinfo{journal}{IEEE Transactions on Consumer Electronics}
  \bibinfo{volume}{69} (\bibinfo{year}{2023}) \bibinfo{pages}{217--225}.
%Type = Article
\bibitem[{Lee et~al.(2022)Lee, Kim, Kim, Choi, Zitouni, Khandoker, Jelinek,
  Hadjileontiadis, Lee, and Jeong}]{r7}
\bibinfo{author}{P.~Lee}, \bibinfo{author}{H.~Kim}, \bibinfo{author}{Y.~Kim},
  \bibinfo{author}{W.~Choi}, \bibinfo{author}{M.~S. Zitouni},
  \bibinfo{author}{A.~Khandoker}, \bibinfo{author}{H.~F. Jelinek},
  \bibinfo{author}{L.~Hadjileontiadis}, \bibinfo{author}{U.~Lee},
  \bibinfo{author}{Y.~Jeong},
\newblock \bibinfo{title}{Beyond pathogen filtration: Possibility of smart
  masks as wearable devices for personal and group health and safety
  management},
\newblock \bibinfo{journal}{JMIR mHealth and uHealth} \bibinfo{volume}{10}
  (\bibinfo{year}{2022}) \bibinfo{pages}{e38614}.
%Type = Article
\bibitem[{Hamada et~al.(2022)Hamada, Yoshida, Kurihara, and Watanabe}]{r8}
\bibinfo{author}{Y.~Hamada}, \bibinfo{author}{T.~Yoshida},
  \bibinfo{author}{Y.~Kurihara}, \bibinfo{author}{K.~Watanabe},
\newblock \bibinfo{title}{Respirometry rate measurement by pyroelectric effect
  of a piezo sounder—monitor and alarm by single device},
\newblock \bibinfo{journal}{IEEE Sensors Journal} \bibinfo{volume}{22}
  (\bibinfo{year}{2022}) \bibinfo{pages}{21197--21208}.
%Type = Article
\bibitem[{Paeng et~al.(2024)Paeng, Shanmugasundaram, We, Kim, Park, Lee, and
  Yim}]{paeng2024rapid}
\bibinfo{author}{C.~Paeng}, \bibinfo{author}{A.~Shanmugasundaram},
  \bibinfo{author}{G.~We}, \bibinfo{author}{T.~Kim}, \bibinfo{author}{J.~Park},
  \bibinfo{author}{D.-W. Lee}, \bibinfo{author}{C.~Yim},
\newblock \bibinfo{title}{Rapid and flexible humidity sensor based on
  laser-induced graphene for monitoring human respiration},
\newblock \bibinfo{journal}{ACS Applied Nano Materials} \bibinfo{volume}{7}
  (\bibinfo{year}{2024}) \bibinfo{pages}{4772--4783}.
%Type = Article
\bibitem[{Pan and Tompkins(1985)}]{4122029}
\bibinfo{author}{J.~Pan}, \bibinfo{author}{W.~J. Tompkins},
\newblock \bibinfo{title}{A real-time qrs detection algorithm},
\newblock \bibinfo{journal}{IEEE Transactions on Biomedical Engineering}
  \bibinfo{volume}{BME-32} (\bibinfo{year}{1985}) \bibinfo{pages}{230--236}.
%Type = Inproceedings
\bibitem[{Mishra and Gupta(2024)}]{10570893}
\bibinfo{author}{R.~Mishra}, \bibinfo{author}{H.~P. Gupta},
\newblock \bibinfo{title}{A federated learning approach to minimize
  communication rounds using noise rectification},
\newblock in: \bibinfo{booktitle}{2024 IEEE Wireless Communications and
  Networking Conference (WCNC)}, \bibinfo{year}{2024}, pp.
  \bibinfo{pages}{1--6}.
%Type = Inproceedings
\bibitem[{Codling et~al.(2024)Codling, Shulkin, Chang, Zhang, Latapie, Noh,
  Zhang, and Dong}]{flohr}
\bibinfo{author}{J.~R. Codling}, \bibinfo{author}{J.~D. Shulkin},
  \bibinfo{author}{Y.-C. Chang}, \bibinfo{author}{J.~Zhang},
  \bibinfo{author}{H.~Latapie}, \bibinfo{author}{H.~Y. Noh},
  \bibinfo{author}{P.~Zhang}, \bibinfo{author}{Y.~Dong},
\newblock \bibinfo{title}{Flohr: Ubiquitous heart rate measurement using
  indirect floor vibration sensing},
\newblock in: \bibinfo{booktitle}{Proceedings of the 11th ACM International
  Conference on Systems for Energy-Efficient Buildings, Cities, and
  Transportation}, \bibinfo{year}{2024}, p. \bibinfo{pages}{44–54}.
%Type = Article
\bibitem[{Sanchez-Vazquez et~al.(2012)Sanchez-Vazquez, Nielen, Gunn, and
  Lewis}]{sanchez2012using}
\bibinfo{author}{M.~J. Sanchez-Vazquez}, \bibinfo{author}{M.~Nielen},
  \bibinfo{author}{G.~J. Gunn}, \bibinfo{author}{F.~I. Lewis},
\newblock \bibinfo{title}{Using seasonal-trend decomposition based on loess
  (stl) to explore temporal patterns of pneumonic lesions in finishing pigs
  slaughtered in england, 2005--2011},
\newblock \bibinfo{journal}{Preventive Veterinary Medicine}
  \bibinfo{volume}{104} (\bibinfo{year}{2012}) \bibinfo{pages}{65--73}.

\end{thebibliography}

\end{document}